\documentclass[letterpaper,twocolumn,10pt]{article}
\usepackage{style/usenix}

\usepackage{color}
\usepackage{multirow}
\usepackage{array}
\usepackage{url}
\usepackage{algorithmicx,algorithm}
\usepackage[noend]{algpseudocode}
\usepackage{enumitem} 
\usepackage{times}
\usepackage{xspace}
\usepackage{latexsym}
\usepackage{amsmath}
\usepackage{booktabs}
\usepackage{tabularx}
\usepackage{balance}
\usepackage{graphicx}
\usepackage{svg}
\usepackage{pifont}
\svgsetup{
    inkscapepath=i/svg-inkscape/
}
\svgpath{{svg/}}
\usepackage{caption}
\usepackage{subcaption}
\usepackage{amssymb}
\usepackage{etoolbox}
\usepackage[all]{nowidow}
\usepackage{authblk}
\usepackage{enumitem}
\setitemize{noitemsep,topsep=0pt,parsep=0pt,partopsep=0pt}
\usepackage{multirow}  
\usepackage{array}  
\usepackage{dblfloatfix}

\setlist{nolistsep}
\newcommand{\parabf}[1]{\medskip\noindent\textbf{#1}}

\newcommand{\paraf}[1]{\noindent\textbf{#1}}
\newcommand{\cut}[1]{}

\newcommand{\ie}{{i.e.}}

\newcommand{\sysname}{MegaScale\xspace}

\newcommand{\revision}[1]{{#1}}

\newcommand{\zhi}[1]{\textcolor{blue}{(Zhi: #1)}}

\begin{document}
\sloppy
\date{}

\definecolor{ao}{rgb}{0.0, 0.5, 0.0}
\definecolor{method}{rgb}{0., 0., 0.}

\title{MegaScale: Scaling Large Language Model Training to More Than 10,000 GPUs}


\author{
\vspace{-0.1in}
\rm{
    Ziheng Jiang$^{\text{1}, *}$ \enskip
    Haibin Lin$^{\text{1}, *}$ \enskip
    Yinmin Zhong$^{\text{2}, *}$ \enskip
    Qi Huang$^{\text{1}}$ \enskip
    Yangrui Chen$^{\text{1}}$ \enskip
    Zhi Zhang$^{\text{1}}$ \enskip\\
}
\vspace{-0.1in}
\rm{
    Yanghua Peng$^{\text{1}}$ \enskip
    Xiang Li$^{\text{1}}$ \enskip
    Cong Xie$^{\text{1}}$ \enskip
    Shibiao Nong$^{\text{1}}$ \enskip
    Yulu Jia$^{\text{1}}$ \enskip
    Sun He$^{\text{1}}$ \enskip
    Hongmin Chen$^{\text{1}}$ \enskip\\
}
\vspace{-0.1in}
\rm{
    Zhihao Bai$^{\text{1}}$ \enskip
    Qi Hou$^{\text{1}}$ \enskip
    Shipeng Yan$^{\text{1}}$ \enskip
    Ding Zhou$^{\text{1}}$ \enskip
    Yiyao Sheng$^{\text{1}}$ \enskip
    Zhuo Jiang$^{\text{1}}$ \enskip\\
}
\vspace{-0.1in}
\rm{
    Haohan Xu$^{\text{1}}$ \enskip
    Haoran Wei$^{\text{1}}$ \enskip
    Zhang Zhang$^{\text{1}}$ \enskip
    Pengfei Nie$^{\text{1}}$ \enskip
    Leqi Zou$^{\text{1}}$ \enskip
    Sida Zhao$^{\text{1}}$ \enskip\\
}
\vspace{-0.1in}
\rm{
    Liang Xiang$^{\text{1}}$ \enskip
    Zherui Liu$^{\text{1}}$ \enskip
    Zhe Li$^{\text{1}}$ \enskip
    Xiaoying Jia$^{\text{1}}$ \enskip
    Jianxi Ye$^{\text{1}}$ \enskip
    Xin Jin$^{\text{2}, \dag}$ \enskip
    Xin Liu$^{\text{1}, \dag}$ \enskip\\
}

{$^{\text{1}}$\textit{ByteDance}\enskip $^{\text{2}}$\textit{Peking University}\enskip}
}

\maketitle
\pagestyle{plain}
\captionsetup[figure]{font=small}
\captionsetup[table]{font=small}

{\let\thefootnote\relax\footnote{{$^*$Equal contribution.}}}
{\let\thefootnote\relax\footnote{{$^\dag$Corresponding authors.}}}
\begin{abstract}

We present the design, implementation and engineering experience in building and
deploying \sysname, a production system for training large language models
(LLMs) at the scale of more than 10,000 GPUs. Training LLMs at this scale brings
unprecedented challenges to training efficiency and stability. We take a
full-stack approach that co-designs the algorithmic and system components across
model block and optimizer design, computation and
communication overlapping, operator optimization, data pipeline, and network performance tuning.
Maintaining high efficiency throughout the training process (i.e.,
stability) is an important consideration in production given the long extent of
LLM training jobs. Many hard stability issues only emerge at large scale, and
\revision{in-depth observability} is the key to address them. We develop a set of diagnosis
tools to monitor system components and events deep in the stack, identify root
causes, and derive effective techniques to achieve fault tolerance and mitigate
stragglers. \sysname achieves 55.2\% Model FLOPs Utilization (MFU) when training
a 175B LLM model on 12,288 GPUs, improving the MFU by 1.34$\times$ compared to
Megatron-LM. We share our operational experience in identifying and fixing
failures and stragglers. We hope by articulating the problems and sharing our
experience from a systems perspective, this work can inspire future LLM systems
research.

\end{abstract}

\section{Introduction}
\label{sec:introduction}

Large language models (LLMs)~\cite{brown2020language} have emerged as a transformative
technology in artificial intelligence (AI). Recent advancements in LLMs have
significantly improved their capability. LLMs have demonstrated tremendous
potential in a wide range of domains, such as machine translation, text
summarization, and conversational agents~\cite{chatgpt}. As a company serving billions
of users, we have been aggressively integrating AI into our products, and we are
putting LLMs as a high priority to shape the future of our products.

Training LLMs is a daunting task that requires enormous computation resources.
The scaling law~\cite{kaplan2020scaling} dictates that the model size and the
training data size are critical factors that determine the model capability. To
achieve state-of-the-art model capability, many efforts have been devoted to
train large models with hundreds of billions or even trillions of parameters on
hundreds of billions or even trillions of tokens. For example, GPT-3~\cite{floridi2020gpt}
has 175 billion parameters and PaLM~\cite{chowdhery2022palm} has 540 billion parameters. Major
players in this field build large-scale AI clusters with tens of thousands of
GPUs to train LLMs.

Scaling LLM training to tens of thousands of GPUs brings unprecedented
challenges. As AI has been at the core of many of our products, we have extensive
experience in training deep neural networks (DNNs). Yet, training a model like
ResNet~\cite{he2016deep} only takes tens or hundreds of GPUs.
Compared to these models,
the scale of training LLMs
is unparallel. While we are not new to building and operating
large-scale GPU clusters, these clusters are normally shared by many training
jobs. Now, in the context of LLM training, a single job is occupying tens of
thousands of GPUs and taking all the resources. The sheer scale of
LLM training introduces two specific challenges from a systems perspective.

The first challenge is to achieve high training efficiency at scale. Model FLOPs
utilization (MFU) is the ratio of the observed throughput to the theoretical
maximum throughput assuming 100\% of peak FLOPs~\cite{narayanan2021efficient}.
It is a standard metric to evaluate training efficiency that directly translates
to end-to-end training speed. LLM training is not embarrassingly parallel. To
train an LLM, the model is split across GPUs and the GPUs heavily communicate
with each other to make progress. Besides communication, other factors such as
operator optimization, data preprocessing and GPU memory consumption also
contribute significantly to MFU.

The second challenge is to achieve high training stability at scale, i.e.,
maintaining high training efficiency throughout the training process. Stability
is particularly important from a production perspective, as LLMs take a long
time to train. Training an LLM with one trillion tokens can take weeks. The
scale and time are orders of magnitude larger than those of regular DNN training
jobs. Failures and stragglers are the norm rather than the exception for LLM
training. At such a scale, the consequences of failures and stragglers are
devastating. Failures are very expensive, and it is critical to reduce the
recovery time, given the large scale. A straggler not only affects its own work,
but slows down the entire job involving tens of thousands of GPUs. 

In this paper, we present the design, implementation and engineering experience
of \sysname, a production system for training LLMs at scale. \sysname enables us
to scale LLM training to more than 10,000 GPUs. We are able to harness the power
of the massive number of GPUs to train LLMs with high training efficiency and
stability. In building and operating \sysname, we apply two systems principles:
algorithm-system co-design and \revision{in-depth observability}. 

\sysname is a specialized system tailored for LLM training. Algorithm-system
co-design is a key principle to maximize performance for specialized systems,
which has been applied widely in computer systems. We apply this
principle to \sysname in the context of LLM training with a full-stack approach
that spans all important system components. We make several modifications and
incorporate effective optimization techniques to the model architecture,
including parallel transformer block~\cite{chowdhery2022palm},
sliding window attention~\cite{beltagy2020longformer} and LAMB optimizer~\cite{You2020Large}. We leverage mixed parallelism strategies that combine data parallelism, pipeline parallelism, tensor
parallelism, and sequence parallelism. Importantly, we design custom techniques based
on the pattern of each parallelism strategy to maximize the overlapping between
communication and computation. We apply prefetching and
tree-based loading to optimize the data pipeline. We leverage non-blocking
asynchronous operations and eliminate global barriers for large-scale collective
communication group initialization. We design a custom network topology, reduce
ECMP hash conflicts, customize congestion control, and tune retransmit timeout
parameters for high network performance.

Stability problems including failures and stragglers in large-scale systems are
notoriously hard to diagnose and fix. Many hard stability issues only emerge at
large scale, which can stem from a wide range of software and hardware faults
deep in the stack. Manually identifying and resolving every single issue is infeasible
given the scale and complexity of the system. \revision{We apply the principle of in-depth observability to build a set of diagnosis tools. By 'in-depth observability', we mean a comprehensive monitoring and visualization strategy that penetrates beyond surface-level metrics to gather detailed, granular data across every component of the system stack, aiming to create a multidimensional view of system performance. The set of tools allows us to diagnose the system and identify root causes, by uncovering the intricate interactions and dependencies that contribute to stability issues.} We develop a robust training framework to automate fault
localization and recovery. We design heartbeat messages encapsulating various
forms of information to facilitate real-time anomaly detection and provide
early warnings. We implement a suite of diagnostic tests to identify nodes
causing disruptions. We optimize the checkpointing and recovery procedure to
reduce interruptions. To troubleshoot nuanced cases caused by stragglers, we develop a performance analysis tool to record fine-grained
CUDA events and generate system-wide heat-map and timeline trace from a distributed view, and develop a 3D parallel
training visualization tool to show data dependencies between ranks for diagnosis.

\sysname is deployed in our datacenters to train LLMs for our products. Over the
years, we have built several AI clusters with different size and hardware
configurations. Our largest AI cluster has over 10,000 GPUs. In terms of
training efficiency, \sysname achieves 55.2\% MFU when training a standard 175B
transformer model on 12,288 GPUs, providing an improvement of 1.34$\times$
compared to the state-of-the-art open-source training framework
Megatron-LM~\cite{shoeybi2020megatronlm}. In terms of model converge and
stability,
we show a real production run of \sysname that trains a proprietary
model with hundreds of billions of parameters on multi-trillion tokens for
several weeks. Over the weeks, the loss continues to converge, and \sysname
repairs and recovers the training process for over 100 times in presence of failures.
We also share our experience in diagnosing and fixing some intriguing problems. 
\revision{We are working on open-sourcing components that can benefit the community on GitHub\footnote{https://github.com/volcengine/veScale}.}

\section{Background}
\label{sec:background}

The training of LLMs, characterized by their vast model architectures and
massive datasets, is computationally intensive. Parallelism strategies
distribute the training process across multiple devices.

\begin{figure}
    \includegraphics[width=0.49\textwidth]{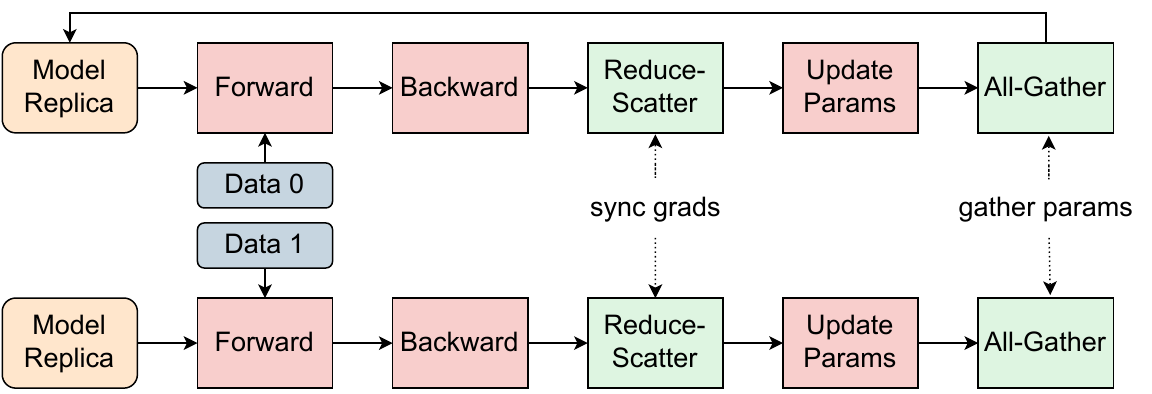}
    \caption{Data parallel training with ZeRO2.}
    \label{fig:dp-zero}
\end{figure}

\parabf{Data parallelism.} It replicates the model and optimizer states across
multiple devices and the data is evenly divided among all devices. Each model
replica executes the forward and backward propagation computation in parallel.
Upon completion of each iteration, all model replicas synchronize to update the
model. Instead of duplicating model states (like the optimizer states,
gradients, and parameters), Zero Redundancy Optimizer (ZeRO) \cite{zero} shards these states
across every data-parallel process. As a result, the traditional all-reduce
operations that aggregate gradients are decomposed into separate reduce-scatter
and all-gather operations. This is because every data-parallel process retains
only a fraction of the total state. ZeRO is structured into three incremental
stages of optimizations. Notably, the second stage is commonly adopted to shard
both the optimizer states and gradients, while ensuring no additional
communication overhead is introduced (Figure~\ref{fig:dp-zero}).

\begin{figure}
    \includegraphics[width=0.46\textwidth]{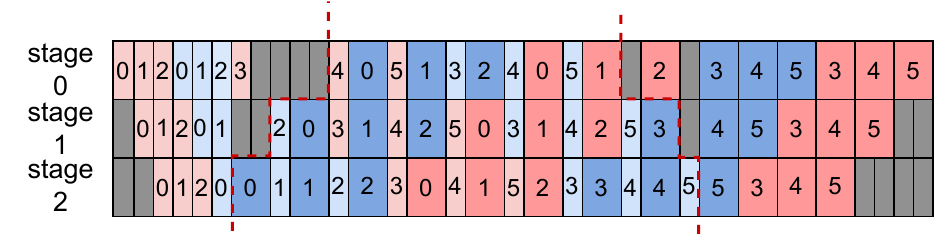}
    \caption{Interleaved 1F1B pipeline.}
    \label{fig:pipeline}
\end{figure}

\parabf{Pipeline parallelism.} It distributes model layers among multiple
devices and each  device owns a portion of the model. Meanwhile, each training
batch is subdivided into a number of micro-batches for pipelined execution. To
reduce pipeline bubbles, various pipeline scheduling strategies are proposed,
e.g., GPipe~\cite{huang2019gpipe}, PipeDream 1F1B~\cite{pipedream}, etc. Megatron-LM~\cite{narayanan2021efficient} employs
the interleaved 1F1B scheduling. Each pipeline stage on every worker is
subdivided into multiple virtual stages, which represents a subset of layers,
referred to as a model chunk. Initially, workers enter a \textit{warm-up} phase,
executing the forward pass for a limited number of in-flight micro-batches.
Following the warm-up, each worker progresses to the \textit{steady} phase where
workers perform one forward pass followed by one backward pass, often
abbreviated as 1F1B. Upon concluding a batch, workers finalize the backward
passes for any remaining in-flight micro-batches during this \textit{cool-down}
phase. Figure~\ref{fig:pipeline} shows an three-stage pipeline where each stage is
further divided into two virtual stages.

\parabf{Tensor parallelism.} It distributes individual operators over multiple
devices, with each device executing a portion of the computation in parallel.
Depending on the specific partitioning strategy and its relationship to prior
and subsequent operators in the model, partitioning can require communication
among participating GPUs to split the input and then merge the output. For example, we
can split GEMMs in the MLP and self-attention blocks among multiple GPUs to
utilize more computational units. Some other operations like LayerNorm and
Dropout are less computationally intensive but demand a considerable amount of
activation memory. Another form of tensor parallelism called sequence parallelism is
proposed to distribute these operators along the sequence dimension to
effectively reduce the activation memory footprint.

\parabf{Combination of parallelism strategies.}
These parallelism strategies can be combined into 3D parallelism to scale the
training of LLMs across many GPUs~\cite{shoeybi2020megatronlm}. Given the high
communication overhead associated with tensor parallelism, it is preferable to
confine such communication within a single cluster node. Conversely, data
parallelism and pipeline parallelism are more amenable to inter-node
communication. In this case, we choose to prioritize building the data
parallelism groups over pipeline parallelism, which can mitigate cross-minipod
communication for data parallelism. 
\section{Efficient Training at Scale}
\label{sec:train}

In the realm of LLMs, efficient training at scale becomes paramount. As we venture into deeper and more expansive models, the computational demands surge explosively. Handling such computation requirements without compromising on model accuracy necessitates the adoption of state-of-the-art algorithmic optimizations, communication strategies, data pipeline management, and network performance tuning techniques. This section delves deep into the methods employed to optimize the training of large models in order to achieve high training efficiency at scale.

\subsection{Algorithmic Optimizations}
\label{sec:algorithm}

We make a few modifications and incorporate recent optimizations at
the algorithmic level to improve training efficiency, without compromising
accuracy. We validate
the impact of these techniques on model convergence in~\S\ref{sec:experience:covergence}.

\parabf{Parallel transformer block~\cite{gpt-j}.} We adopt a parallel version of the transformer block in lieu of the standard serialized formulation. 
Specifically, the standard formula of the transformer block can be reformatted from 
\begin{equation}
    \begin{aligned}
        y = x + \text{MLP}(\text{LN}(x + \text{Attention}(\text{LN}(x))))
    \end{aligned}
\end{equation}
into
\begin{equation}
    \begin{aligned}
        y = x + \text{MLP}(\text{LN}(x)) + \text{Attention}(\text{LN}(x))
    \end{aligned}
\end{equation}

With this approach, the computation of the attention block and the MLP block can
be executed in parallel, thereby reducing the computation time. Prior
work~\cite{chowdhery2022palm} shows that this modification does not degrade the
quality of models with parameters in the hundreds of billions.

\begin{figure*}[t!]
    \centering
     \includegraphics[width=0.96\textwidth]{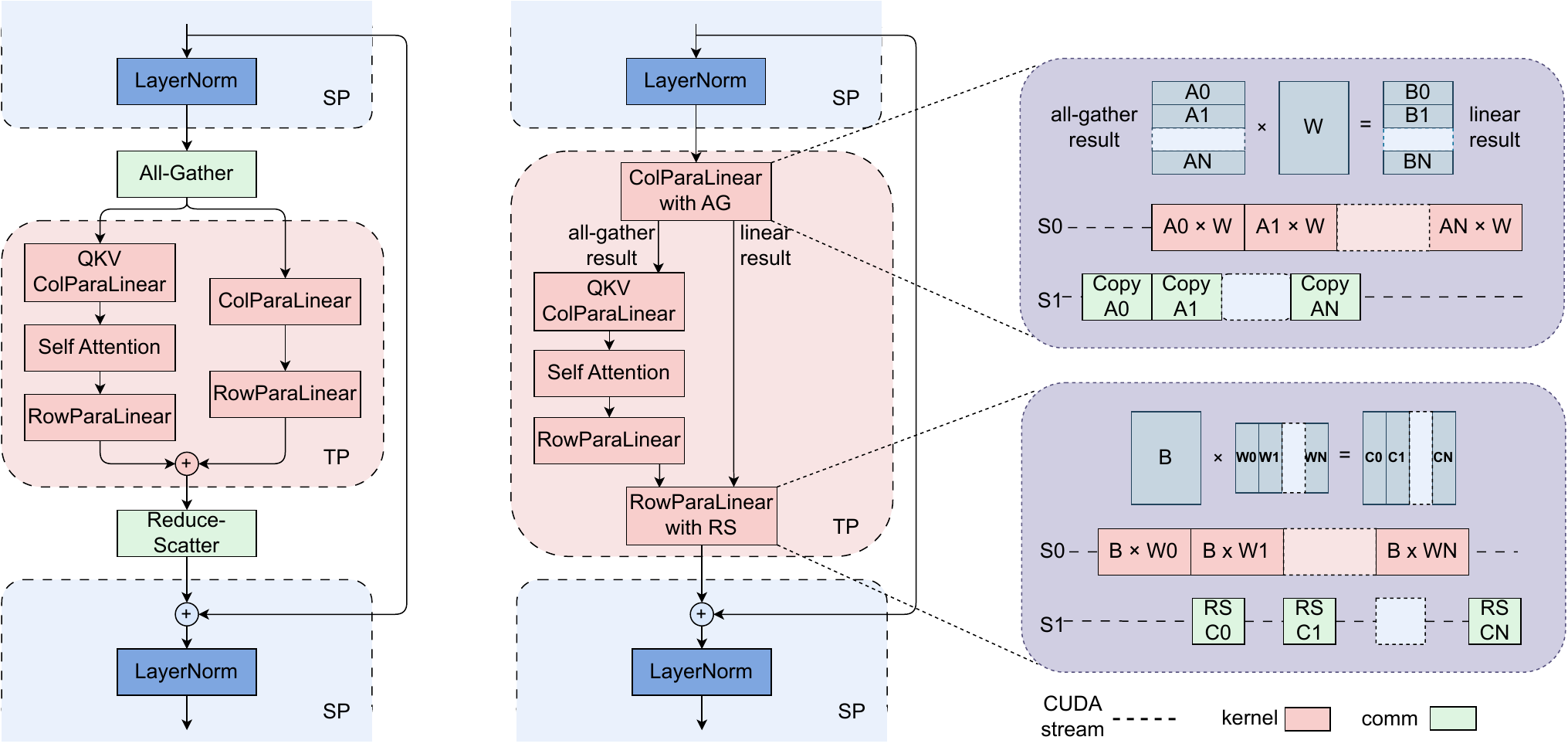}
     \\
     (a) PTB with SP and TP \hspace*{0.6in} (b) Fuse communication into Linears \hspace*{0.4in} (c) Overlap communication with GEMM
     \caption{Overlapping communication in tensor parallelism (TP) and sequence parallelism (SP) with parallel transformer block (PTB).}
    \label{fig:ptb}
\end{figure*}

\parabf{Sliding window attention (SWA)}. Sliding window
attention~\cite{beltagy2020longformer} is a sparse attention mechanism that
employs a fixed-size window surrounding each token in the input sequence. The
computation complexity is $O(s \times w)$, where $s$ is the input sequence
length and $w$ is the fixed window size. Sliding window attention is more
efficient than the full self-attention, whose computation complexity is $O(s
\times s)$, given that $w\ll s$. \revision{Past work~\cite{beltagy2020longformer} and our micro-benchmark (\S\ref{sec:experience:covergence}) have shown that} 
the information across the entire input can be retained with a large receptive
field created by stacking layers of such windowed attention. This enables faster training without compromising the accuracy.

\parabf{LAMB optimizer}. Efficient training at a large scale is often
hindered by batch size constraints. Particularly, increasing the batch size may
adversely affect model convergence. The LAMB optimizer~\cite{You2020Large} has been
demonstrated to enable the scaling of BERT's training batch size to 64K without
compromising accuracy. In the LLM setting, our experiments find that LAMB
can scale the batch size to 4$\times$ without accuracy loss. With
interleaved pipeline parallelism, the original schedule contains $\frac{4}{v}
\frac{p-1}{m}$ pipeline bubbles when training four steps with 1$\times$ batch
size~\cite{narayanan2021efficient}, while the pipeline bubbles of training one
step with 4$\times$ batch size are $\frac{1}{v} \frac{p-1}{4m}$. Hence, {\sysname}
reduces 87.5\% of the pipeline bubbles via LAMB optimizer.

\subsection{Communication Overlapping in 3D Parallelism}
\label{sec:overlap}

To reduce the iteration time, we systematically analyze the dependencies between
computation and communication for all the operators in 3D parallelism, and
design techniques to hide the overhead of all the off-the-critical-path
operations.

\parabf{Overlapping in data parallelism.} 
As shown in Figure~\ref{fig:dp-zero}, for data parallelism, two main
communication operations stand out. One is the \textit{all-gather} operation,
which fetches the most recent model parameters from workers in other data
parallel ranks during the forward pass. The other is the \textit{reduce-scatter}
operation, which collect the gradients in the backward pass. In 3D parallelism,
a single device may host multiple model chunks. Overlapping is implemented on a
model chunk basis to maximize bandwidth utilization. The \textit{all-gather}
operation is triggered prior to the forward pass of a model chunk, and the
\textit{reduce-scatter} operation commences after its backward pass. This
results in a challenge where the first \textit{all-gather} operation and the
last \textit{reduce-scatter} operation cannot be hidden. Inspired by PyTorch
FSDP~\cite{pytorch_fsdp}, the initial \textit{all-gather} operation is
pre-fetched at the beginning of each iteration, allowing it to overlap with data
loading operations, effectively reducing the communication time by a factor of
$1/(2*vpp\_size)$.
We also launch the high priority communication first to maximize overlapping. The priorities of communication operators are determined by the order of the corresponding computation operators that depend on the communication result.

\begin{figure}[t]
    \centering
     \includegraphics[width=0.46\textwidth]{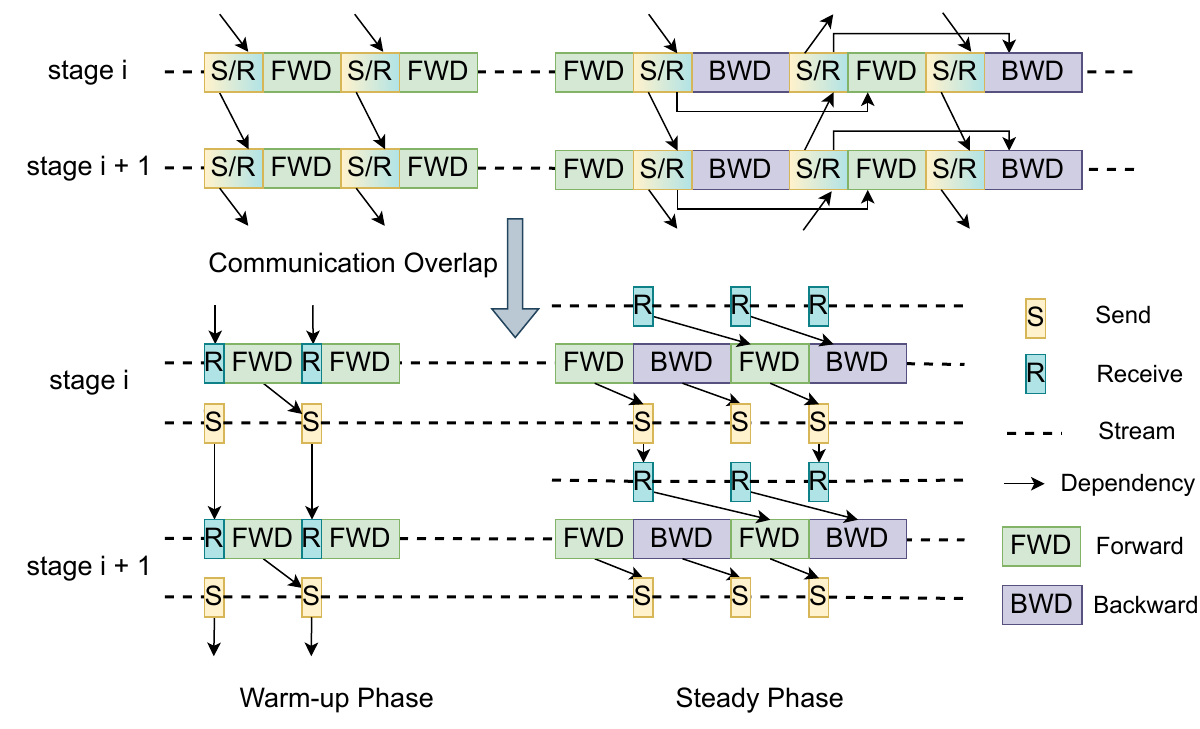}
     \hfill
     \caption{Overlapping communication in pipeline parallelism.}
    \label{fig:pp-overlap}
\end{figure}

\parabf{Overlapping in pipeline parallelism.} Pipeline parallelism features point-to-point \textit{send/receive} communication. \sysname uses the interleaved 1F1B scheduling method mentioned in~\ref{sec:background}. We note that in the warm-up phase, the forward pass only depends on its previous \textit{receive}. We thus decouple the \textit{send} and \textit{receive}, which are often implemented together and can be blocked by the slower one. By breaking this dependency, we enable the send operation to overlap with the computation as shown in the left part of Figure~\ref{fig:pp-overlap}. The cool-down phase can be viewed as the inverse of the warm-up phase, allowing for the inverse application of the same technique. As for the steady phase, both the forward and backward computation are independent of adjacent communication operations. Taking the backward as an example, as shown in the right part of Figure~\ref{fig:pp-overlap}, its previous \textit{receive} is for the next forward computation while the \textit{send} is for the backward computation in the previous stage. So the \textit{send} and \textit{receive} operations can be launched asynchronously to overlap with the computation.

\parabf{Overlapping in tensor/sequence parallelism.} Tensor parallelism is commonly used to partition weights in computational-intensive operations, while operations like \textit{LayerNorm} and \textit{Dropout} are partitioned along the sequence dimension to save GPU memory. This necessitates \textit{all-gather} and \textit{reduce-scatter} operations for input collection and output redistribution across GPUs. Figure~\ref{fig:ptb}\textcolor{green!80!black}{a} shows this communication pattern in the parallel transformer block architecture. Here the two communication operators are in the critical path. To eliminate this overhead, we choose to fuse \textit{all-gather} and \textit{reduce-scatter} with the parallel \textit{Linears} on the FFN path (Figure \ref{fig:ptb}\textcolor{green!80!black}{b}). Since the GEMM kernels on the FFN
path is larger, the communication can be hidden better. We break the GEMM kernel into small chunks, and pipeline the execution with the communication (Figure~\ref{fig:ptb}\textcolor{green!80!black}{c}). 
This strategy can be applied in the backward pass similarly.

\subsection{Efficient Operators}

Despite the optimization for GEMM operators in Megatron-LM, we identify
opportunities for further enhancement in other operators. For the attention part,
we adopt FlashAttention-2~\cite{dao2023flashattention}, which improves
work partitioning between different thread blocks and warps. For LayerNorm and GeLU, we observe that they
are composed of fine-grained kernels in previous implementations. By fusing
these kernels together, we reduce the overhead associated with launching
multiple kernels and aid in optimizing memory access patterns, thereby
achieving better performance.

\subsection{Data Pipeline}
\label{sec:data}

Data preprocessing and loading are often overlooked. However, these operations
create non-negligible GPU idle time at the beginning of each training step.
Optimizing these operations are essential for efficiency of the training
process.

\parabf{Asynchronous data preprocessing.} Data preprocessing is not on the
critical path. As a result, while the GPU workers are synchronizing gradients at
the end of each training step, the data preprocessing for the subsequent step
can start, which hides the preprocessing overhead.

\parabf{Redundant dataloader elimination.} In a typical data loading phase of
distributed training, each GPU worker is equipped with its own data loader,
responsible for reading training data into the CPU memory before forwarding it
to the GPU. This leads to competition among workers for disk read bandwidth,
thereby creating a bottleneck. Notably, we observe that in the LLM training
setting, GPU workers within the same machine are in the same tensor parallel
group. Consequently, their inputs for each iteration are inherently identical.
Based on this observation, we adopt a two-layer tree-based approach.
We use a single, dedicated data loader on each machine to read the
training data into a piece of shared memory. Subsequently, each GPU worker is
responsible for copying the necessary data to its own GPU memory. This
eliminates redundant reads and significantly enhances the efficiency of data
transfer.

\subsection{Collective Communication Group Initialization}

In distributed training, the initialization phase involves the establishment of
NVIDIA Collective Communications Library (NCCL) communication groups among GPU
workers. Since this overhead is relatively negligible in small-scale scenarios,
\texttt{torch.distributed} is used by default. As the number of GPUs scales to
over ten thousand, the overhead introduced by naive implementations becomes
intolerable. \revision{We conduct experiments on the same AI cluster in \S\ref{sec:experience} and our empirical measurement indicates that the initialization time for Megatron-LM on 2,048 NVIDIA Ampere GPUs is approximately 1047 seconds.} While this may
appear relatively small compared to the training duration, it imposes a
significant hurdle to routine testing and iterative development (e.g., minor
code adjustments in hyperparameter tuning and debugging). It also hampers
the implementation of fast restart-and-recovery mechanisms.

\revision{To address this issue, we perform a detailed profiling of
\texttt{torch.distributed}~\cite{li2020pytorch}} and identify two primary causes of excessive
initialization time. The first issue resides in the synchronization step, where
each process is involved in a barrier operation at the end of initialization a
specific communication group. \revision{This barrier uses TCPStore, an inner distributed Key-Value Store implementation in Pytorch which
operates in a single-threaded, blocking read-write manner.} We replace TCPStore
with Redis, which is non-blocking and asynchronous. This reduces the
initialization time to 361 seconds on 2,048 GPUs. The second issue is related to
the incautious usage of global barriers. Each process executes a global barrier
after initializing its corresponding communication group. We carefully design
the order in which communication groups are initialized to minimize the need for
global barriers. This approach lowers the time complexity of the global barrier from $O(n^2)$ to $O(n)$. The initialization time is reduced to under 5 seconds on 2048 GPUs, and to under 30 seconds on more than 10,000 GPUs with those optimizations.

\subsection{Network Performance Tuning}

We analyze the traffic across machines in 3D parallelism and design techniques to improve network performance.

\parabf{Network topology.} Our datacenter network is built with high-performance
switches based on Broadcom Tomahawk 4 chips. The total bandwidth of each Tomahawk
chip is 25.6Tbps with 64$\times$400Gbps ports. Three layers of switches are
connected in a CLOS-like topology to connect more than 10,000 GPUs. For switches
at each layer, the bandwidth percentage between downlink and uplink is 1:1. That
is, 32 ports are used as downlink and 32 ports are used as uplink. The network
provides high bandwidth with a small diameter. Every node can communicate with
other nodes within a limited number of hops.   

\parabf{Reducing ECMP hashing conflicts.}
We carefully design the network topology and schedule network traffic to reduce ECMP hashing conflicts. First, at the
top-of-rack (ToR) switch level, one 400G downlink port is split into two 200G
downlink ports with specific AOC cables. The conflict
probability is reduced as the bandwidth of each uplink is double of that
of a downlink. Second, eight 200G NICs on the server
is connected to eight different switches in a multi-rail way. The number of GPU
servers connected by the same sets of ToR switches can reach 64. And we strategically schedule the data-intensive nodes from our training tasks to operate under the same Top of Rack (ToR) switch. This approach significantly reduces the number of switch hops required for communication and further reduce ECMP hashing conflicts probability.

\parabf{Congestion control.} \revision{In distributed training, all-to-all communication may lead to congestion and elevated levels of Priority Flow Control (PFC)\cite{PFCStandard} when employing the default DCQCN\cite{zhu2015congestion} protocol at scale. Excessive use of PFC can result in head-of-line (HoL) blocking \cite{zhu2015congestion}, thereby diminishing network throughput. To mitigate these issues, we have developed an algorithm incorporating principles from both Swift\cite{kumar2020swift} and DCQCN, which integrates the precise measurement of Round-Trip Time (RTT) with the rapid congestion response capabilities of Explicit Congestion Notification (ECN). This approach significantly enhances throughput and minimizes congestion related to PFC.}

\parabf{Retransmit timeout setting.} Parameters in NCCL can be set to control
retransmit timer and retry count. We tune these parameters for fast recovery under link
flapping.
To further reduce the recover time, we enable the adap\_retrans
feature on the NIC. This feature enables retransmission in a shorter interval
and help recover the transmission more quickly when the link flapping period is short.
\section{Fault Tolerance}
\label{sec:fault_tolerance}

As the training cluster scales to over tens of thousands of GPUs, software and
hardware faults become virtually inevitable. We introduce a robust training
framework for LLM training that achieves automatic fault identification
and fast recovery, enabling fault tolerance with minimal human intervention and
negligible impact on ongoing training tasks.

\begin{figure}
    \includegraphics[width=0.45\textwidth]{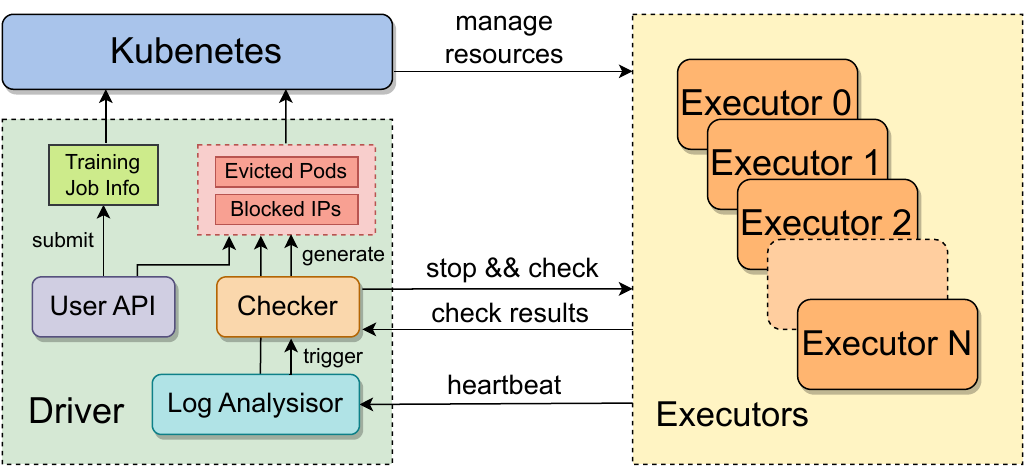}
    \caption{Robust training workflow.}
    \label{fig:robust_training_workflow}
\end{figure}

\subsection{Robust Training Workflow}
\label{sec:robust_training_workflow}

As Figure~\ref{fig:robust_training_workflow} shows, upon receiving a submitted
training task, the driver process interfaces with a custom Kubernetes to
allocate computing resources and initiate the corresponding Pod for each
executor. One executor manage one node. Once the executor has completed a
series of initialization tasks, it creates the training process on each GPU and
a robust training daemon which sends heartbeat to the driver periodically. These
heartbeats encapsulate various forms of information to enable real-time
anomaly detection and issue early warnings (\S\ref{sec:log}). When the driver
process detects an abnormal status in a particular training process, or fails to
receive a heartbeat from an executor within a predefined time window, it
triggers the fault recovery procedure. The driver will suspend the ongoing
training task across all executors and command them to run a series of
self-check diagnostics (\S\ref{sec:faults_checker}). These diagnostic tests are
carefully designed to be lightweight yet comprehensive, covering the majority of
common hardware and software faults. Once the problematic nodes are identified,
the driver submits the IP addresses of the nodes to be blocked, along with the
information of the Pods running on them, to Kubernetes, which evicts the faulty
nodes and replenishes the cluster with an equivalent amount of healthy ones
which pass our diagnostic tests. Additionally, we provide a user interface that
allows for manual eviction of nodes, particularly for those identified through
manual analysis as in~\S\ref{sec:monitor}. After the recovery process is
complete, the driver resumes training from the latest checkpoint. We optimize
the checkpoint and resume process to minimize the loss of training progress
(\S\ref{sec:checkpoint}). 

\subsection{Data Collection and Analysis}
\label{sec:log}

The heartbeat messages includes the basic information of the executor, such as
the IP address, the Pod name, and hardware information, etc. Additionally, the
current status of the training processes is reported, enabling the driver to
promptly detect any explicit anomalies. The \textit{stdout/stderr} logs of
training processes are also included. They will be aggregated, filtered and
analyzed on the fly.
If specific warning or error keywords are detected, the driver
will report real-time diagnostic information. Moreover, RDMA traffic metrics
are also included, serving as an indicator for network utilization and efficiency.
Some anomalies in the training process may not manifest as explicit errors,
giving the appearance that training is proceeding as expected. In such cases,
RDMA traffic metrics serve as a critical indicator. Given the periodic nature of
the training tasks, the network traffic characteristics for each step should
exhibit similar patterns. Therefore, any significant decline or abnormal
fluctuation in RDMA traffic is a signal of potential anomalies. Upon detecting
such irregularities, the driver will issue alerts for manual investigation. If
the traffic ceases entirely, the driver will automatically initiate the fault
recovery procedure. 

In order to enhance the monitoring of training stability and performance, we have developed a monitoring system with precision reaching the millisecond level. Different levels of monitoring are employed to track various indicators. Second-level monitoring is typically used to assess the overall health status and to rule out common configuration impacts on training. For instance, ECN/PFC/QoS configurations, link flapping, or any other issues of NICs. Millisecond-level monitoring, on the other hand, is used to determine if the network is congested and whether the data transfer speed of data parallelism and pipe parallelism has reached its physical limit.

\subsection{Diagnostic Tests}
\label{sec:faults_checker}

There exists a trade-off between execution time and accuracy in self-check
diagnostics. Extended diagnostic duration can adversely affect the effective
training time, while high false positive rates can lead to unnecessary
exclusion of machines that are actually functional. Through iterative
experimentation and optimization, we have deployed a suite of lightweight
diagnostic tests that effectively cover a broad spectrum of hardware and
software faults encountered during actual training processes.

\parabf{Intra-host network tests.} To diagnose potential bottlenecks in
intra-host network, we use our internally developed tool to test two things. The
\textit{Loopback} test measures the loopback bandwidth from all RDMA NICs
(RNICs) to various intra-host endpoints, including memory nodes and GPUs. It
conducts a full-mesh test within the host, covering all possible link
combinations. This allows us to infer link-specific bandwidth degradation and
irregularities in PCIe configurations based on end-to-end bandwidth results. The
second \textit{RNIC-to-RNIC} test examines the connectivity and bandwidth
performance between different RNICs on the same host. These tests provide
insights into whether the RNICs meet the hardware speed specifications and
whether the underlying routing configurations are correctly configured.

\parabf{NCCL tests.} To identify potential faults in GPU communication, we run
an \textit{all-to-all} test among the GPUs within a single node to observe
whether the bandwidth aligns with expected benchmarks. Once intra-host
communication test is passed, each node also conducts an \textit{all-reduce}
test with neighboring machines under the same ToR switch to assess inter-node
GPU communication.

\subsection{Fast Checkpointing and Recovery}
\label{sec:checkpoint}

After identifying and evicting faulty machines, the driver needs to resume the
training by loading model weights and optimizer states from the most recent
checkpoint. It is critical to ensure that the latest
checkpoint is as close as possible to the state of training progress when the
faults happened, to minimize loss in computation and time. This require us
to increase the frequency of checkpointing during training. However, we also
want to reduce the latency introduced by the checkpointing process, especially
the time on the critical path which blocks the training progress, thus impeding
the overall system throughput.

To achieve fast checkpointing, we introduce an optimized, two-stage approach. In
the first stage, each GPU worker writes its on-chip states to the host memory,
and then continues the training process. After the optimization of Pytorch's
serialization mechanism and the use of pinned memory, this process can be
reduced to several seconds thanks to the high PCIe bandwidth, thereby minimally
interrupting the ongoing training process. In the second stage, a background
process takes over, asynchronously transferring the state from the host memory
to a distributed file system (HDFS in our deployment) for centralized maintenance. This decoupling of operations into two
stages allows the GPU workers to resume training almost immediately after
dumping their state, while the more time-consuming process of writing to HDFS is
offloaded to a separate, non-blocking process.

In the context of recovery from a checkpoint, it is on the critical path since
training can not be started without the last checkpoint. The bottleneck is the
bandwidth of HDFS, especially when each GPU worker needs to read its
corresponding state partition. To alleviate this bottleneck, we propose an
optimized data retrieval strategy. We recognize that multiple GPU workers often
share the same state partition, e.g., the workers in the same data parallel
group. Accordingly, we designate a single worker in the group to read the shared
state partition from HDFS, thereby reducing the load linearly. This worker then
broadcasts the state partition to all other GPU workers that share the same
data. This approach effectively mitigates the bandwidth constraints of HDFS,
leading to a substantial reduction in the recovery time.

\begin{figure}[t]
    \includegraphics[width=0.46\textwidth]{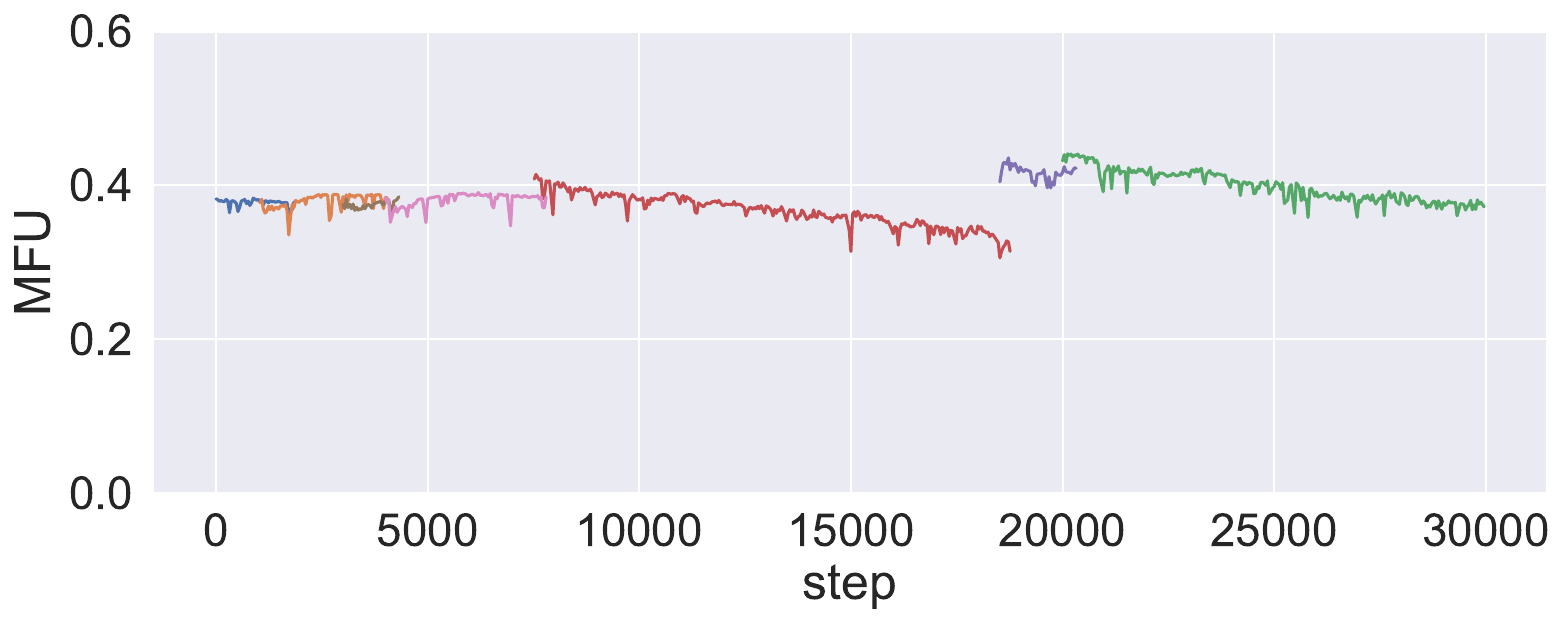}
    \caption{Inconsistent MFU observed in large-scale training. Different colors denote distinct executions of the same training job.}
    \label{fig:mfu-straggler}
\end{figure}

\section{Training Troubleshooting}
\label{sec:monitor}

Although our robust training framework automatically discovers, pinpoints, and
resolves the majority of common faults, there remain certain hardware anomalies
that manifest probabilistically and cannot be found by machine self-checks. Some
anomalies may make the system appear to operate normally, yet significantly
degrades the training efficiency.
To address these nuanced cases, we have
implemented several custom monitoring and analysis tools designed to support
case-by-case anomaly detection.

\subsection{Performance Diagnosis with CUDA Event Monitor}

At the scale of tens of thousands of GPUs, we observe that, unlike in
smaller-scale experiments, different runs exhibit varying computational
efficiencies. Even with identical configurations , this inconsistency persist, as shown in Figure \ref{fig:mfu-straggler}. We also observed that the performance of training tasks is not consistent at this scale. The MFU for various training tasks gradually declines over time. While this leads us to suspect variations between individual machines, no evident variations are detected under single GPU GEMM micro-benchmarks. To diagnose those performance issues, we develop a performance analysis tool that records the execution
time of critical code segments on each machine rank during a run. In contrast to
previous tools such as the torch profiler or the Megatron-LM timer, our tool times
events based on the CUDA events method. This approach minimizes the need for
CUDA synchronization, thus preventing performance degradation, allowing us to
consistently run it in our production training jobs. This tool offers two
visualization modes and can analyze the collected data from different
perspectives.

\begin{figure}[t!]
    \includegraphics[width=0.45\textwidth]{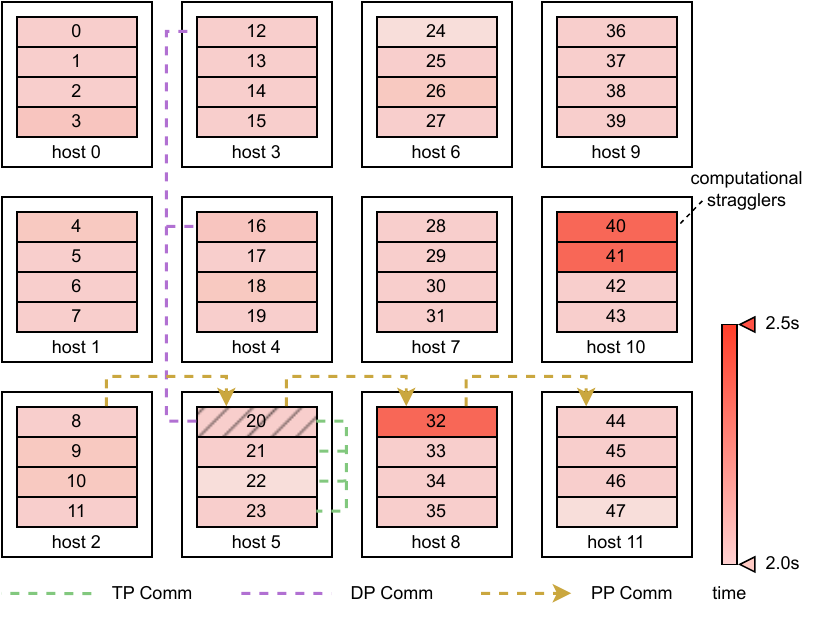}
    \caption{Performance heat-map. The color denotes the running time of the code segments on a rank. The figure also shows the 3D visualization feature, where rank 20 has been selected and the dependency across different parallelism dimensions become visible.}
    \label{fig:heatmap}
\end{figure}

The first mode uses a heat map to show time consumption differences between
machines from various dimensions, depicted in Figure \ref{fig:heatmap}. We gather latency data of the computation
phase (forward and backward) across devices and average the latency across
steps. The aggregated data is visualized using a heat-map. The heat-map reveals
that a minor fraction of machines (approximately 0.5\%) exhibit substantially
slower performance during training, thereby hindering overall training progress.
The training efficiency is predominantly determined by the slowest machine's
performance (\ie, stragglers), leading to inconsistencies in training efficiency
across diverse runs, since machine scheduling within the cluster is stochastic.
After excluding these outlier machines, the peak MFU across runs becomes
consistent.

\begin{figure*}
    \centering
    \includegraphics[width=0.95\textwidth]{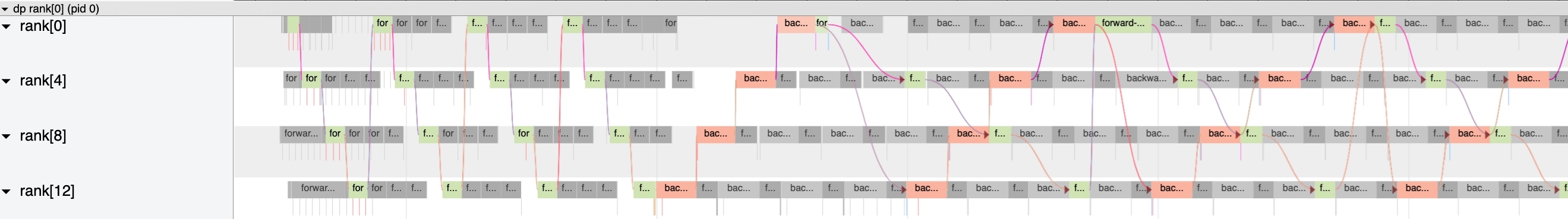}
    \caption{The trace shows events collected in a pipeline group on a unified timeline. Dependencies become visible when an event is selected.}
    \label{fig:trace}
\end{figure*}

The other mode displays the event timeline on machines in a trace format from different distributed views (data parallelism, pipeline parallelism, tensor parallelism). Traditional profiler, such as PyTorch Profiler, is primarily designed for single-node activity analysis. This approach offers limited insight in distributed training scenarios where execution dependencies frequently span across multiple nodes. By aggregating the trace spans of various ranks onto a singular timeline, we gain a comprehensive perspective, revealing the overall execution order, pipeline bubbles, and synchronization characteristics among data parallel ranks. Figure \ref{fig:trace} displays how our distributed tracer visualizes the actual execution of pipeline parallelism, detailing the data dependencies between different pipeline stages through the consolidation of event data across a pipeline parallelism group.

Every piece of data from the CUDA event timer is stored in a remote analytical database, allowing for easy retrieval of details from any step event. While the timer data is wrote to a local file in a line-by-line format, a separate streamer process then synchronizes this log file with a Kafka queue in real-time. The analytical database remains updated by consuming data from this Kafka queue, enabling on-the-fly analysis without interrupting the training job. \revision{All the monitoring features are turned on during real production training and the overhead is negligible compared to the training time.}

\subsection{3D Parallel Training Visualization}
\label{sec:visualization}

With 3D parallelism and our optimization techniques (\S\ref{sec:train}), the landscape of data flow and task sequencing is
exceedingly intricate. Each GPU worker may be engaged in several synchronous
or asynchronous operations at the given moment, leading to complex
dependencies among them. This intricacy amplifies the challenges of fault
diagnosis: when a single GPU worker experiences a fault, the entire cluster of
nodes can stall in the NCCL communication operations, ultimately leading to a
system-wide timeout. Externally, this situation manifests as a generic blockage,
but the root cause of which is often buried under a deluge of timeout messages.
To rapidly pinpoint the problematic nodes, we let each GPU worker log its own
ongoing event upon communication timeout. These logs are then used to construct
a visual representation of data dependencies based on the logical topology in
the 3D parallel setting.

As Figure~\ref{fig:heatmap} shows, the cluster in 3D parallel training can
logically be split into three dimensions: tensor parallelism, pipeline
parallelism, and data parallelism. When we select a specific GPU worker, it
displays its position within the logical topology, the direction of data flow
and the different communication operations it involves. Importantly, in the
event of an error, the tool provides direct access to the worker's error
messages if any. This serves as a powerful tool for diagnosing training
anomalies, enabling quicker identification and resolution of faults.

Consider the aforementioned case when defective GPUs probabilistically cause blocking when executing NCCL communication operations. Such blocking can hang the entire machine, leading to cascading timeouts across other dependent nodes and ultimately resulting in the paralysis of the entire training process. To swiftly identify these faulty nodes, we utilize the 3D parallel training visualization tool. Nodes that timeout due to waiting for the faulty ones will log their ongoing operations upon exiting. In contrast, the nodes with the faulty GPUs are hung and do not log any such information. Therefore, by examining the logs and the data flow within the visualization, these problematic nodes can be easily pinpointed. Once identified, these nodes can be manually isolated and flagged for maintenance through the robust training framework, as described in \ref{sec:robust_training_workflow}.
\section{Experience}
\label{sec:experience}

In this section, we describe our deployment and operational experience of
\sysname. We build dedicated AI clusters for LLM training. Over the years, we
have iterated several versions of our specialized AI cluster architecture, and
we are currently operating several AI clusters with varying size and hardware
configurations. We use these AI clusters to train a wide range of models, from
computer vision and recommendation models to LLMs. With the increasing
importance of LLMs, we are building AI clusters with larger size to cater the
need of LLM training. As of September 2023, the largest AI cluster in our
production for LLM training contains more than 10,000 NVIDIA Ampere GPUs. We are
also in the process of building large clusters based on the newest NVIDIA Hopper
GPUs, as NVIDIA is ramping up production.

\begin{table}[!t]
    \def\barr{\begin{tabular}{c}}
    \def\earr{\end{tabular}}
    \centering
    \begin{tabular}{c|c|c|c|c|c}
        \hline
         \barr {Model}\\{Size} \earr &
         {Heads} & 
         \barr {Hidden}\\{Size} \earr &
         {Layers} & {TP} & {PP}  \\
         \hline
         175B & 128 & 12288 & 96 & 8 & 8 \\
         530B & 160 & 20480 & 105 & 8 & 35 \\
         \hline
    \end{tabular}
    \caption{Model configurations.}
    \label{tab:exp-model-config}
\end{table}

\begin{table*}[ht]
    \centering
    \begin{tabular}{|c|c|c|c|c|c|c|c|}
        \hline
        \multirow{2}{*}{Batch Size} & \multirow{2}{*}{Method} & \multirow{2}{*}{GPUs} & \multirow{2}{*}{\shortstack{Iteration Time (s)}} & \multirow{2}{*}{\shortstack{Throughput \\ (tokens/s)} } &  \multirow{2}{*}{\shortstack{Training Time \\(days)}} & \multirow{2}{*}{MFU} & \multirow{2}{*}{\shortstack{Aggregate \\PFlops/s }} \\
                                &                           &       & & & & & \\
        \hline
        \multirow{8}{*}{768}    & \multirow{4}{*}{Megatron-LM} &  256  & 40.0 & 39.3k  & 88.35 & 53.0\% & 43.3 \\
                                &                           &  512  & 21.2 & 74.1k  & 46.86 & 49.9\% & 77.6 \\
                                &                           &  768  & 15.2 & 103.8k & 33.45 & 46.7\% & 111.9 \\
                                &                           & 1024  & 11.9 & 132.7k & 26.17 & 44.7\% & 131.9 \\
        \cline{2-8}
                                & \multirow{4}{*}{\sysname} & 256  & 32.0 & 49.0k  & 70.86 & 65.3\%(\textbf{1.23$\times$}) & 52.2 \\
                                &                           & 512  & 16.5 & 95.1k  & 36.51 & 63.5\%(\textbf{1.27$\times$}) & 101.4 \\
                                &                           & 768  & 11.5 & 136.7k & 25.40 & 61.3\%(\textbf{1.31$\times$}) & 146.9 \\
                                &                           & 1024 & 8.9  & 176.9k & 19.62 & 59.0\%(\textbf{1.32$\times$}) & 188.5 \\
        \hline
        \multirow{8}{*}{6144}   & \multirow{4}{*}{Megatron-LM} & 3072  & 29.02 & 433.6k  & 8.01 & 48.7\% & 466.8 \\
                                &                           & 6144  & 14.78 & 851.6k  & 4.08 & 47.8\% & 916.3 \\
                                &                           & 8192  & 12.24 & 1027.9k & 3.38 & 43.3\% & 1106.7 \\
                                &                           & 12288 & 8.57  & 1466.8k & 2.37 & 41.2\% & 1579.5 \\
        \cline{2-8}
                                & \multirow{4}{*}{\sysname} & 3072  & 23.66 & 531.9k  & 6.53 & 59.1\%(\textbf{1.21$\times$}) & 566.5 \\
                                &                           & 6144  & 12.21 & 1030.9k & 3.37 & 57.3\%(\textbf{1.19$\times$}) & 1098.4 \\
                                &                           & 8192  & 9.56  & 1315.6k & 2.64 & 54.9\%(\textbf{1.26$\times$}) & 1400.6 \\
                                &                           & 12288 & 6.34  & 1984.0k & \textbf{1.75} & 55.2\%(\textbf{1.34$\times$}) & 2166.3 \\
        \hline
    \end{tabular}
    \caption{Strong-scaling training performance for the 175B model. We set the batch size to 6144 when training with 3072 to 12288 GPUs. For 256 to 1024 GPUs, we decrease the batch size to 768 due to GPU memory limit. We report the training time required for training 300B tokens here. The number in parentheses in the MFU column represents the speedup of \sysname compared to Megatron-LM.}
    \label{tab:exp-175b-scaling}
\end{table*}

\subsection{Training Performance}

\revision{\sysname is built on top of Megatron-LM~\cite{narayanan2021efficient}, which is a state-of-the-art
open-source LLM training framework that integrates 3D parallelism techniques and
takes advantage of hardware resources.} Our experiments use the \revision{Megatron-LM (commit hash: 285068c8) on Github~\cite{megatrongithub}, chosen for its stability and feature set at the commencement of our experiments months ago.
} We use the same batch size for Megatron-LM and
\sysname for fair comparison. We use two model sizes: 175B parameters and 530B
parameters. We use interleaved pipeline-parallel
schedule~\cite{korthikanti2023reducing} with six and three interleaving stages
for the 175B and 530B models, respectively. Sequence length is 2,048 and
vocabulary size is 64,000 for all the cases. Table~\ref{tab:exp-model-config}
shows the details of the model configuration.

\begin{figure}[!t]
    \includegraphics[width=0.46\textwidth]{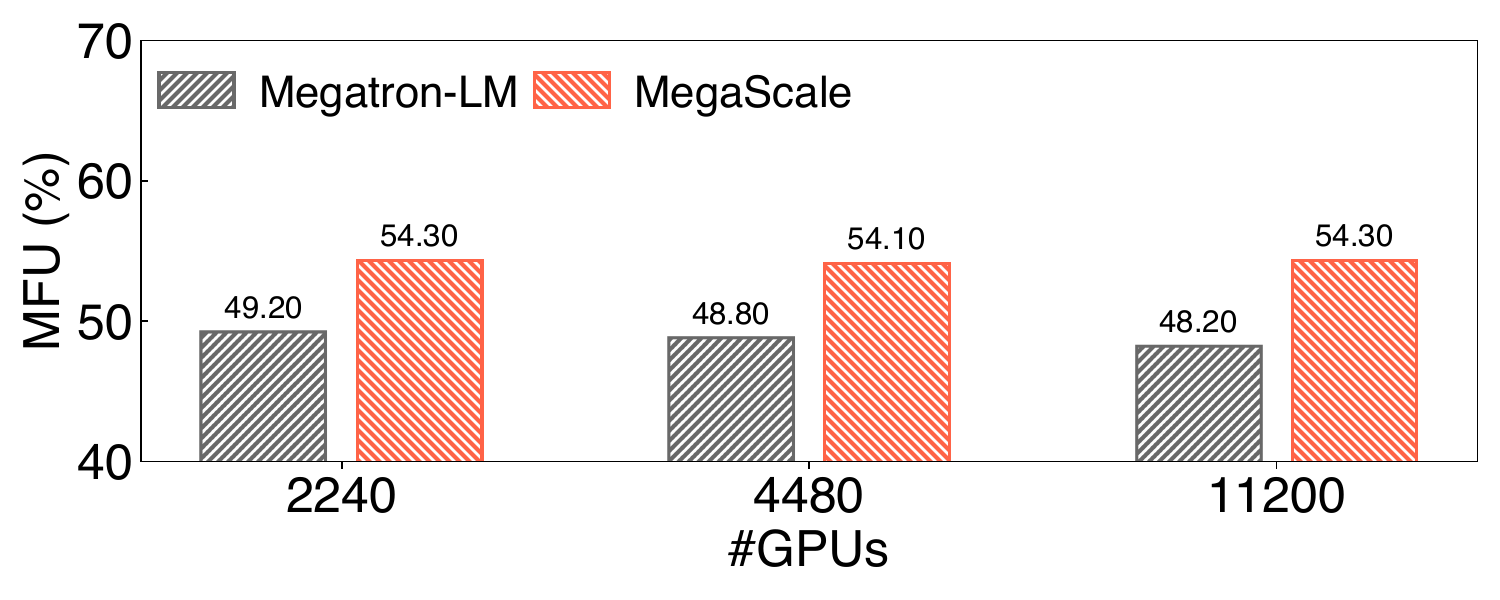}
    \caption{Weak-scaling training performance of Megatron-LM and \sysname on the 530B model, where the batch size is scaled proportionally with the number of GPUs.}
    \label{fig:exp_530b_scaling}
\end{figure}
\parabf{Scalability.} Figure~\ref{fig:exp_530b_scaling} compares Megatron-LM
and {\sysname} when training the 530B model, where we set
the batch size as the number of GPUs with adjusted learning rate to show the MFU results. We see that the MFU of {\sysname} is higher than Megatron-LM
by up to 6.1\%. With increasing scales, the MFU of Megatron-LM decreases by 1.6\%
with more stragglers and communication, while {\sysname} has near-linear
scalability due to 3D-parallel communication overlapping. 

In Table \ref{tab:exp-175b-scaling}, we evaluate the strong-scaling training performance of Megatron-LM and {\sysname} on the 175B model by increasing number of GPUs and maintaining a constant batch size. This experimental setting is more realistic, given that batch size is constrained by convergence effects and cannot be indefinitely scaled with the number of GPUs. {\sysname} achieves up to 1.34$\times$ speedups over Megatron-LM across all settings. With increasing GPUs, we observe the MFU of {\sysname} decreases from 59.1\% to 55.2\%. This is expected since the batch size is fixed and the computation-to-communication ratio decreases with more GPUs. Even in the largest scale with 12,288 GPUs, {\sysname} still outperforms Megatron-LM by 14\% MFU. For the smaller scale training, the speedup of {\sysname} over the baseline ranges from 1.23$\times$ to 1.32$\times$. Note that the difference in the maximum number of GPUs between this and the previous experiments (e.g., 12,288 vs. 11,200) is due to distinct 3D parallelism configurations for 175B and 530B models.

\parabf{Ablation study.} 
We evaluate the effectiveness of our optimization techniques of {\sysname}.
Table~\ref{tab:exp-ablation} shows the MFU improvement breakdown with different
optimizations when training the 175B model on 256 GPUs. The baseline \revision{is the original Megatron-LM and} has 47.7\%
MFU. It is worth noting that the networking optimizations are 
turned on for both Megatron-LM and {\sysname} in this evaluation. We first apply 
two algorithmic techniques, parallel transformer block and
sliding window attention, to Megatron-LM, achieving 5.6\% MFU improvement. 
Communication is the major bottleneck of large-scale LLM training, and
the 3D parallel communication overlapping of {\sysname} hides the overhead and
accelerates training by 6.2\% MFU. We further adopt efficient operators and obtain 
1.7\% acceleration. Other optimizations such as data pipeline 
optimizations and the problematic code elimination mentioned 
in \ref{sec:faults_fixed} further achieves 1.1\%
performance gain. Finally, we scale the batch size from 256 to 768 with LAMB
optimizer, which significantly extends the steady phase in interleaved pipeline
parallelism and achieves 3.0\% MFU improvement.  To sum up, {\sysname}
outperforms the baseline by 17.6\% in the MFU number with all these
optimizations.

\begin{table}[t]
    \centering
    \begin{tabular}{|c|c|c|}
        \hline
        Idx & Method & MFU ($\Delta$ MFU) \\
        \hline
        1 &  baseline & 47.7\% \\
        2 &  (1) with PTB & 52.3\% (4.6\%)  \\
        3 &  (2) with SWA & 53.3\% (5.6\%) \\
        4 &  (3) with TP overlap & 55.5\% (7.8\%) \\
        5 &  (4) with PP overlap &  58.0\% (10.3\%)\\
        6 &  (5) with DP overlap &  59.5\% (11.8\%)\\
        7 &  (6) with efficient operators &  61.2\% (13.5\%)\\
        8 &  (7) with misc optimizations & 62.3\% (14.6\%)  \\
        9 & (8) with LAMB (BS$\times$3) &  65.3\% (17.6\%)\\
         \hline
    \end{tabular}
    \caption{MFU improvement breakdown when training the 175B model with 256 GPUs and batch size 256.}
    \label{tab:exp-ablation}
\end{table}

\subsection{Model Convergence and Stability}
\label{sec:experience:covergence}

\begin{figure*}[t]
     \centering
     \begin{subfigure}[b]{0.46\textwidth}
         \centering
         \includegraphics[width=\textwidth]{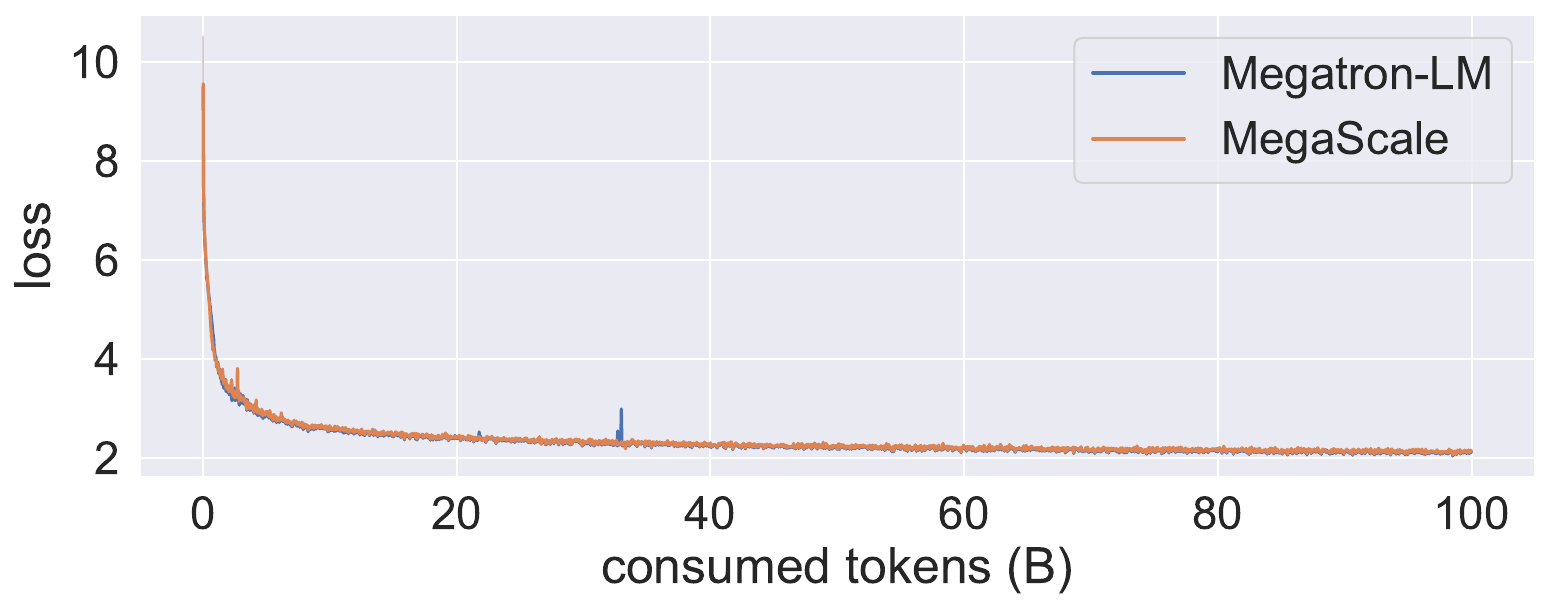}
         \caption{The training loss curve of {\sysname}, which includes algorithm optimizations, in comparison with Megatron-LM.}
         \label{fig:algo-loss}
     \end{subfigure}
     \hfill
     \begin{subfigure}[b]{0.46\textwidth}
         \centering
         \includegraphics[width=\textwidth]{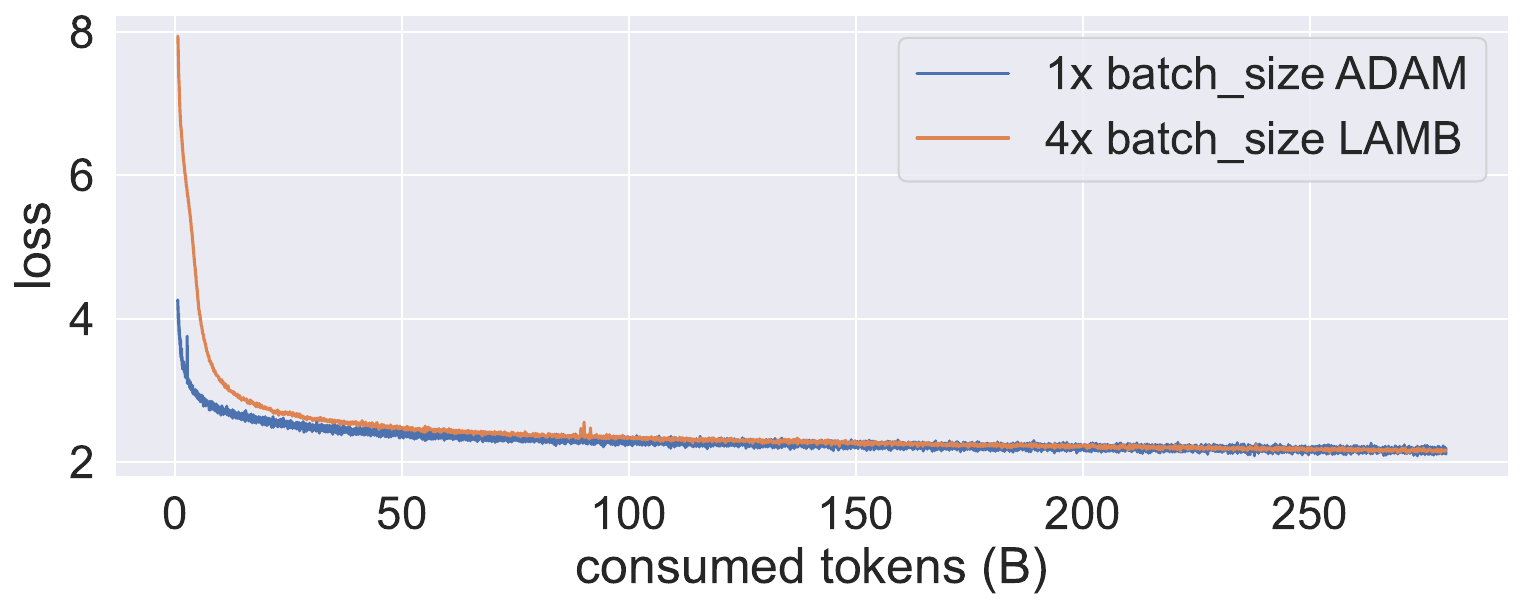}
         \caption{The training loss curve of ADAM optimizer and LAMB optimizer with four times of the batch size.}
         \label{fig:lamb-loss}
     \end{subfigure}
    \caption{The training loss curves in microbenchmark experiments.}
    \label{fig:loss}
\end{figure*}

\begin{figure*}[t]
    \centering
    \includegraphics[width=0.95\textwidth]{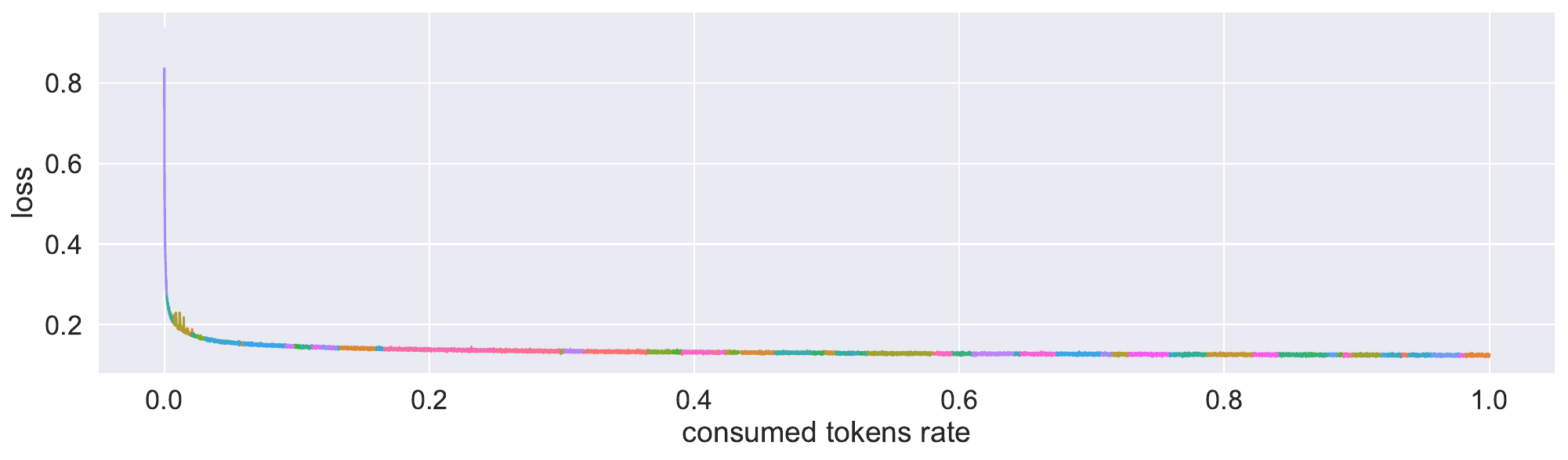}
    \caption{The normalized training loss curve of a real production run on more than
    10,000 GPUs for several weeks. This run trains a model with hundreds of
    billions of parameters on multi-trillion tokens. Different colors indicate
    training restarts. \sysname repairs and recovers the training process for
    over 100 times in presence of failures.}
    \label{fig:llm-loss}
\end{figure*}
\paraf{Model convergence microbenchmarks.}
 \revision{We first conduct microbenchmark experiments to validate the algorithm techniques do not affect the model convergence. Due to the resource limit, the microbenchmarks are done on the 13B model.}
As shown in Figure~\ref{fig:algo-loss}, while {\sysname} adopts algorithm techniques, including parallel transformer 
block and sliding window attention, it achieves comparable loss results with 
the baseline when training with more than 100B tokens. We also evaluate the 
effect of LAMB optimizer as depicted in Figure~\ref{fig:lamb-loss}, which shows 
that LAMB optimizer with four times of batch size achieves the same loss as ADAM 
optimizer after around 250B tokens. \revision{Based on these observations, we turn on all the algorithmic optimizations in production training.}

\parabf{Model convergence and stability in real production LLM training.}
We show the model convergence and stability from a real production run. This run
trains a proprietary model with hundreds of billions of parameters on
multi-trillion tokens. This run uses more than 10,000 GPUs and lasts
for several weeks. Figure~\ref{fig:llm-loss} shows the loss continues to
converge, with distinct colors indicating the training is restarted. Over the
several weeks of this run, we experience training restarts over 100 times. With
the robust training framework, over 90\% of software and hardware faults are automatically identified and fixed by the techniques
detailed in~\S\ref{sec:fault_tolerance}. The rest of the problems are handled
with the help of the troubleshooting tools described in~\S\ref{sec:monitor}.

\subsection{Problems Discovered and Fixed}
\label{sec:faults_fixed}

\revision{We conduct an analysis of the fault records for the aforementioned production training job over several weeks. Our findings indicate that over 90\% of the exceptions among them are automatically detected, located, and recovered using our robust training framework, such as CUDA error and segmentation fault. The average time required for detecting failure and executing diagnostic tests is less than 10 minutes. Moreover, the system can catch up to the training progress prior to the crash within 15 minutes from the latest checkpoints, maintaining over 90\% effective training time rate, which is calculated as the number of iterations multiplied by the iteration training time, divided by the total training time. Below we show our experience in diagnosing and fixing some intriguing problems, which need to be analyzed using the troubleshooting tools in~\S\ref{sec:monitor}.}

\parabf{Computational stragglers.} Building upon our utilization of CUDA event timers, we made another pertinent observation across multiple experimental setups. We noted that specific hosts took approximately 10\% more time to execute the same forward computations compared to other ranks. This consistency across different experiments led us to conclude that the issue was not with the software but rather inherent to certain machines in the cluster. After isolating and removing these problematic hosts from the cluster, we observed an approximate 0.7\% improvement in MFU.

\begin{figure}
    \includegraphics[width=0.46\textwidth]{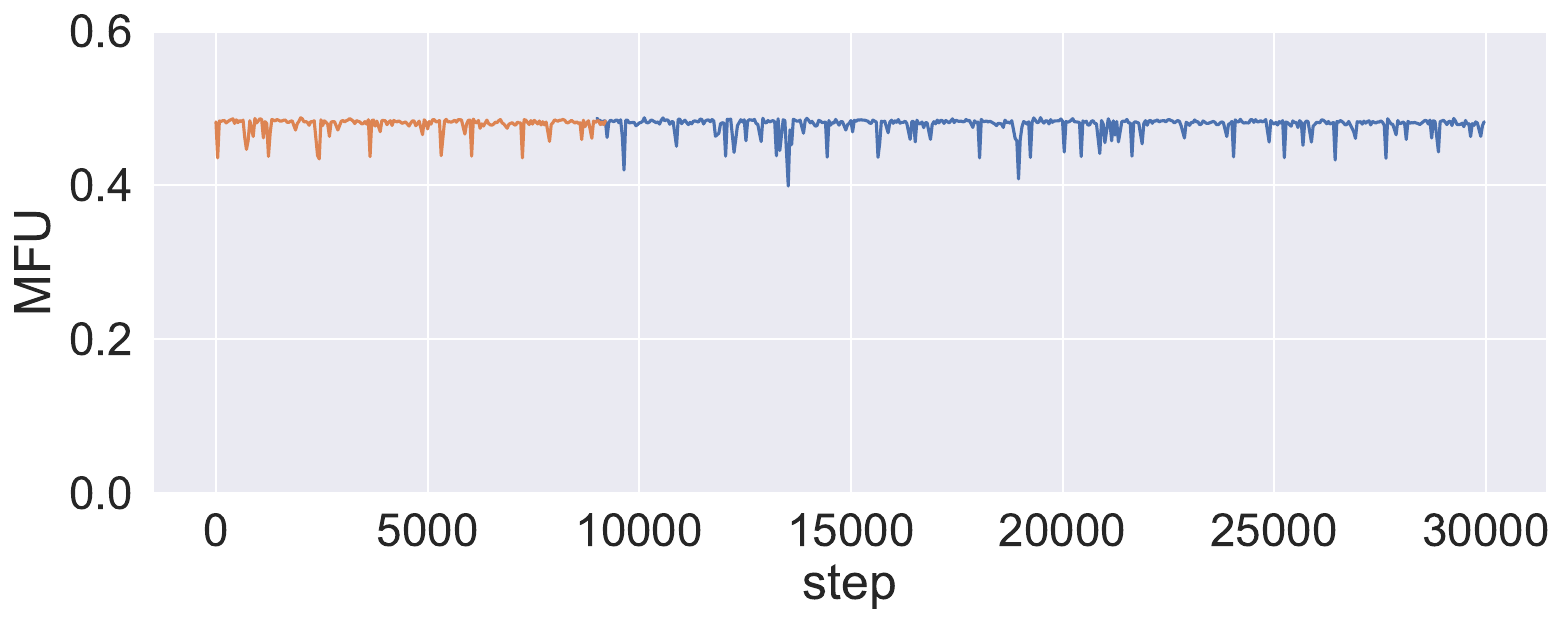}
    \caption{The MFU becomes stable after addressing the stragglers and problematic code segments. \revision{Different colors represent different training trials with the same setup.}}
    \label{fig:mfu-stable}
\end{figure}

\parabf{MFU decreasing.} In such large-scale training experiments, another phenomenon we observed is that training efficiency did not remain consistent over time. Instead, as the training progressed, the MFU of our training job gradually decreased. Through a step-by-step analysis based on CUDA event timer metrics, we noted several key findings. While the time consumed per training step was increasing, the time spent on forward, backward, and optimizer computations remained stable, irrespective of the increasing number of steps. This led us to infer that the time increase must be attributed to the collective communication overhead. Upon a reverse chronological examination, we identified the last collective communication step as the gradient reduce-scatter in data parallelism. If this step is delayed, the overall time per step elongates. Since we observed network bandwidth to be largely stable, we ruled out slowed communication speed as a factor for the increased time. According to the synchronization characteristics of collective communication, this leaves us with one conclusion: some ranks initiate the reduce-scatter operation later than others, forcing a wait for the slowest rank to catch up. In a scaled-down experiment involving only two ranks per data parallel group, we measured the launch times for reduce-scatter calls and found them to not be consistently staggered but rather fluctuating reciprocally. Furthermore, the size of this time stagger increased as more steps were executed. Specifically, Rank A may initially lag behind Rank B but might eventually surpass Rank B in speed and by a growing margin. Ultimately, all ranks waited for the slowest rank. To trace back the root cause of this time skew, we located the variance to occur during the forward computation stage. Digging deeper into the code, we attributed this irregularity to fluctuations caused by 
some code segments. \revision{For instance, irregular garbage collection can introduce disturbances into the training procedure, and certain PyTorch operations can lead to performance fluctuations. } These operations are on the critical path but can be affected along the training procedure. After modifying or removing those problematic code segments, we no longer observed a significant decline in MFU, as shown in Figure~\ref{fig:mfu-stable}.

\parabf{Frequent network interface flapping problem.} We occasionally encounter training stall or training speed drop problem due to frequent network interface flapping. When the network interface flapping phenomena happens, the network interface goes down at first then goes up again. The interval between down and up time usually lasts for several seconds. During the down process, all the packets in transmission will be dropped. The first lesson we learn is the timeout threshold should be set explicitly to a larger value , otherwise the default value will make NCCL timeout very quickly and return a completion error before the network card up again. The second lesson we learn is that the root cause of this problem is the bad link quality between network card, AOC cable and switch. The flapping frequency can be reduced to a satisfactory level by doing lower level quality control over network card signal strength, AOC cable quality and switch side signal strength. 

\section{Related Work}
\label{sec:related}

\paraf{LLM training.}
A lot of efforts have been put to the training of pre-trained LLMs, including
proprietary ones such as GPT-3~\cite{brown2020language},
GPT-4~\cite{openai2023gpt4}, GShard~\cite{lepikhin2020gshard},
PaLM~\cite{chowdhery2022palm}, and many others~\cite{askell2021general,
wei2022finetuned, smith2022using, hoffmann2022training, su2023welm}, as well as
open-source alternatives like OPT~\cite{zhang2022opt},
BLOOM~\cite{scao2022bloom}, Llama~\cite{touvron2023llama1},
Llama-2~\cite{touvron2023llama}. Existing technical reports in the field
predominantly focus on model performance comparisons, leaving out the specific
details of the system infrastructure that makes such training possible.
This paper fills this gap by sharing our experience of end-to-end LLM
pre-training at the scale of over 10,000 GPUs from a systems perspective.

After pre-training, pre-trained base models can be further fine-tuned to adapt
to downstream tasks better. This has led to the emergence of a range of dialogue
models~\cite{alpaca, vicuna2023, koala_blogpost_2023, BELLE} exemplified by
ChatGPT. However, it is worth noting that the computational power and data
requirements for fine-tuning are substantially lower than those for
pre-training. With the application of optimization techniques such as
quantization~\cite{li2020train, xiao2022smoothquant, frantar2022gptq,
dettmers2022llm} and low-rank adaptation~\cite{hu2021lora}, fine-tuning can be
efficiently accomplished with limited resources. 

\parabf{LLM optimizations.}
In addition to the techniques mentioned previously in the paper, there exists a lot of other works targeted at improving the efficiency of LLMs. Sparse or linear attentions~\cite{child2019generating, katharopoulos2020transformers, zhu2021longshort} are proposed to make the memory consumption scales approximately linearly. Several studies
aim to design new architectures rather than conventional
transformer architectures to address the efficiency issue, such as
RWKV~\cite{peng2023rwkv} and RetNet~\cite{sun2023retentive}. 
Many recent studies have been devoted to developing communication acceleration techniques for LLMs. Some works reduce communication traffic using gradient compression~\cite{alistarh2017qsgd} or mixed-precision training~\cite{DBLP:conf/iclr/MicikeviciusNAD18}, while others schedule communication to overlap it with computation. 
Many popular ML frameworks, such as TensorFlow~\cite{abadi2016tensorflow} and PyTorch~\cite{paszke2019pytorch}, enable overlapping communication with backward propagation by default.
Recent works~\cite{jayarajan2019priority, hashemi2019tictac, peng2019generic, bao2020preemptive} further overlap gradient synchronization with forward computation via tensor partitioning, at the cost of extra overhead. Some works~\cite{li2018pipe, chen2022sapipe} introduce fixed staleness to the training pipeline for full overlapping communication and communication. However, the staleness may degrade the model performance.

\parabf{Diagnosis tools in datacenters.} Many diagnosis tools have been developed
to identify and pinpoint hardware and software problems in datacenters.
Pingmesh~\cite{pingmesh} is an active
probing system based on end hosts. Network wide RTT and packet loss and measured
by sending probing ping packets and doing data analysis. Network-wide SLAs are
provided and network problems including packet-blackhole and packet silent drop
are detected.  EverFlow~\cite{everflow}, LossRadar~\cite{lossradar},
NetBouncer~\cite{netbouncer} exploits the capability of switches to diagnose
detailed network problems like network path failures or specific network port
failures. NetBouncer leverages IP-in-IP tunnel techniques to do path probing.
EverFlow requires mirroring network packets to a centralized server to do
debugging. Hostping~\cite{hostping} is a diagnosis system based on
end hosts that focuses on intra-host bottlenecks. It actively senses complex GPU
server PCIe/NVLINK interconnects and does loopback bandwidth and latency tests.

\parabf{Fault tolerance in large-scale distributed systems.}
Fault tolerance has been a major concern in large-scale distributed systems,
where a wide range of hardware and software failures can occur.
Many fault tolerance techniques have been proposed in the past
that cater the needs of different
systems and deployment scenarios.
Reactive fault
tolerance techniques are used to reduce the impact of failures on a system when
the failures occur. There are many techniques in this category such
as Retry~\cite{haider2011fault}, Replication~\cite{haider2011fault},
Checkpointing~\cite{peng2018optimus} and Message
Logging~\cite{tanenbaum2007distributed}.
These techniques incur some system overhead to recover from failures.
Proactive fault tolerance techniques
keep healthy components in place as backups of the faulty
components, obviating the need of recovery from faults and errors, e.g., preemptive
migration~\cite{chakravorty2006proactive, chakravorty2005proactive,
chen2020elastic} and load balancing~\cite{behera2014performance}. However, these
approaches often assume that failures are predictable,
while it is challenging for real
large-scale distributed systems to predict the failures due to the complexity
of the systems.

\section{Conclusion}
\label{sec:conclusion}

In this paper, we offer an in-depth look at the design, implementation and
deployment of \sysname, a production-grade system built to train LLMs at the scale of over 10,000 GPUs. \sysname exploits algorithm-system co-design to optimize training efficiency.
\sysname achieves 55.2\% MFU when training a 175B LLM model on 12,288 GPUs, a 1.34$\times$ improvement over Megatron-LM.
We emphasize the
need for fault tolerance throughout the training process and implement a
tailored robust training framework to locate and fix faults automatically. We
provide a comprehensive set of monitoring tools for deep observability into system
components and events, facilitating root cause identification for intricate
anomalies. We believe that our work not only offers practical insights for those
working on LLM training, but also paves the way for future research in this
rapidly evolving field.

\section*{Acknowledgments}

\revision{We thank our shepherd Deepak Narayanan and the anonymous NSDI
reviewers for their insightful and constructive feedback. We thank Zhenyuan Yang and Wenjia Zhu
for their thorough feedback on earlier drafts of this manuscript. We thank Zuquan
Song, Gaohong Liu and the broader ByteDance Applied Machine Learning team for
their support throughout this project.}
\label{lastpage}

{
\bibliographystyle{ieeetr}
\bibliography{paper}}

\clearpage

\end{document}